\documentclass[letterpaper]{article} 
\usepackage{aaai23}  
\usepackage{times}  
\usepackage{helvet}  
\usepackage{courier}  
\usepackage[hyphens]{url}  
\usepackage{graphicx} 
\usepackage{natbib}  
\usepackage{caption} 
\usepackage{algorithm}
\usepackage{algorithmic}

\usepackage{newfloat}
\usepackage{listings}

\usepackage{bm}
\usepackage{amsmath}
\usepackage{amsthm}

\newtheorem{thm}{Theorem}
\newtheorem{lem}[thm]{Lemma}
\DeclareCaptionStyle{ruled}{labelfont=normalfont,labelsep=colon,strut=off} 
\floatstyle{ruled}
\newfloat{listing}{tb}{lst}{}
\floatname{listing}{Listing}
\title{Nearest-Neighbor Sampling Based Conditional Independence
Testing}
\author{
   Shuai Li\textsuperscript{\rm 1}, Ziqi Chen\textsuperscript{\rm 1}\thanks{Corresponding author: Ziqi Chen.}, Hongtu Zhu\textsuperscript{\rm 2}, Christina Dan Wang\textsuperscript{\rm 3}, Wang Wen\textsuperscript{\rm 4}
}
\affiliations{
    \textsuperscript{\rm 1}School of Statistics, KLATASDS-MOE, East China Normal University, Shanghai, China\\
  \textsuperscript{\rm 2}  Departments of Biostatistics, Statistics, Computer Science, and Genetics,
 The University of North Carolina at Chapel Hill, Chapel Hill, USA\\
\textsuperscript{\rm 3} Business Division, New York University Shanghai, Shanghai,  China\\
\textsuperscript{\rm 4} School of Mathematics and Statistics, Central South University, Changsha, China

    shuaili.cq@foxmail.com, zqchen@fem.ecnu.edu.cn, htzhu@email.unc.edu, christina.wang@nyu.edu, wangwen.a@foxmail.com
}

\usepackage{bibentry}

\begin{document}

\maketitle

\begin{abstract}
The conditional randomization
test (CRT) was recently proposed to test   whether two random variables $X$ and $Y$ are conditionally independent given random variables
 $Z$.
The CRT assumes that the conditional distribution of $X$ given $Z$ is known under the null hypothesis and then it is compared to the distribution of
the observed samples of the original data. The aim of this paper is to
 develop a novel  alternative  of  CRT   by using  nearest-neighbor sampling without assuming  the exact form of the distribution of $X$ given $Z$. Specifically, we  utilize the computationally efficient 1-nearest-neighbor to approximate the conditional distribution that encodes the null hypothesis. Then, theoretically, we show that the distribution of  the generated samples is very close to the true conditional distribution in terms of  total variation distance.  Furthermore, we take the classifier-based conditional mutual information estimator as our test statistic. The test statistic as an empirical fundamental information theoretic quantity  is  able to well capture  the conditional-dependence feature. We show that our proposed  test    is computationally very fast, while controlling type I and II errors quite well. Finally, we demonstrate the efficiency of our proposed test in both synthetic and real data analyses.
\end{abstract}

\section{Introduction}

Conditional independence testing (CIT)  has wide applications  in  statistics and machine learning, including causal inference \cite{spirtes2000causation,pearl2009causality,cai2022distribution}  and graphical models \cite{lauritzen1996graphical,koller2009probabilistic} as two well-known examples. The aim of  this paper is to develop a flexible and fast method for CIT. Specifically, we consider two univariate continuous random variables
$X$ and $Y$, and a set of  random variables  $Z\in R^{d_Z}$, whose dimension $d_Z$ can
potentially diverge to infinity,
with a joint  density function given  by $p_{X,Y,Z}(x,y,z)$.
Based on
$n$ independently and identically distributed (i.i.d)  copies $\{(X_i,Y_i,Z_i): i=1, \ldots, n\} $ of $(X, Y, Z)$,
we are interested in testing
whether two random variables $X$ and $Y$ are conditionally independent given  $Z$;
that is,
\begin{equation*}
H_0: X \perp \!\!\! \perp Y|Z ~~\mbox{versus}~~ H_1:X \not \! \perp \!\!\! \perp Y|Z,
\end{equation*}
where $\perp \!\!\! \perp$ denotes the independence.
  The high dimensionality of    $Z$  makes CIT challenging \cite{bellot2019conditional,shi2021double}. 
Our proposed method can be readily extended
to the scenario of multivariate $X$ and $Y$.

Recently, many  methods have been proposed to test conditional independence. See, for example,  \citeauthor{candes2018panning} \shortcite{candes2018panning},  \citeauthor{zhang2011kernel} \shortcite{zhang2011kernel},  \citeauthor{zhang2017causal} \shortcite{zhang2017causal}, \citeauthor{bellot2019conditional} \shortcite{bellot2019conditional},   \citeauthor{strobl2019approximate} \shortcite{strobl2019approximate},  \citeauthor{berrett2020conditional} \shortcite{berrett2020conditional},
 \citeauthor{shah2020hardness} \shortcite{shah2020hardness}, \citeauthor{shi2021double} \shortcite{shi2021double},  and \citeauthor{zhang2022residual} \shortcite{zhang2022residual}. Among them, the
conditional randomization test (CRT) proposed by \citeauthor{candes2018panning} \shortcite{candes2018panning} is one of the most important methods, but CRT assumes that
the true conditional distribution  $p_{X|Z}$ is known.  Conditional on $\{Z_1, \ldots, Z_n\}$, one can independently draw $X_{i}^{(m)}\sim p_{X|Z=Z_i}$ for each $i$ across $m=1, \ldots, M$ such that  all  $\bm{X}^{(m)}:=(X_{1}^{(m)}, \ldots, X_{n}^{(m)})$ are independent of
$\bm{X}:=(X_1, \ldots, X_n)$ and $\bm{Y}:=(Y_1, \ldots ,Y_n)$, where $M$ is the number of repetitions.   Thus,
under the null hypothesis $H_0:X\perp \!\!\! \perp Y|Z$, we have   $(X^{(m)}, Y, Z)\overset{d}{=}(X, Y, Z)$ for all $m$,
where $\overset{d}{=}$ denotes equality in distribution.
A large difference between $(X^{(m)}, Y, Z)$ and $(X, Y, Z)$  can be regarded as a strong evidence against
  $H_0$.  Statistically, one can
consider a  test statistic $T(\bm X,\bm Y,\bm Z)$ and approximate its  $p$-value by
\begin{equation}\label{CRT}
    \frac{1+\sum_{m=1}^{M}{I}(T(\bm X^{(m)},\bm Y,\bm Z)\geq T(\bm X,\bm Y,\bm Z))} {1+M},
\end{equation}
where ${I}(\cdot)$ is the indicator function. Under $H_0$, the $p$-value is valid and  $P(p\leq \alpha |H_{0})\leq \alpha $ holds for  any $\alpha \in (0,1)$.

Several  methods have been developed based on different approximations to
$p_{X|Z}$, since $p_{X|Z}$ is rarely known in practice.  
For instance, \citeauthor{bellot2019conditional} \shortcite{bellot2019conditional}  developed a  Generative Conditional Independence Test (GCIT) by using Wasserstein generative adversarial networks (WGANs, \citeauthor{arjovsky2017wasserstein}, \citeyear{arjovsky2017wasserstein}) to approximate $p_{X|Z}$.  Let $\widehat{p}_{X|Z}$ be an estimator of ${p}_{X|Z}$ based on WGANs.
Theoretically, as shown in \citeauthor{bellot2019conditional} \shortcite{bellot2019conditional},   the excess type I error over a desired level $\alpha$ of their GCIT test is bounded by $E\{d_{TV}(p_{\bm X|\bm Z}, \widehat{p}_{\bm X|\bm Z})\}$, where $d_{TV}$ denotes the total variation distance.   However, Figure \ref{fig11} shows that $\widehat{p}_{ X| Z}$  approximates $p_{X|Z}$  very poorly  in two relatively simple simulation settings. Thus, as shown in  synthetic data analysis,  the GCIT test has
 inflated  type-I errors. 
 Recently,
 \citeauthor{shi2021double}  \shortcite{shi2021double}   proposed to use the Sinkhorn GANs \cite{genevay2018learning} to approximate $p_{X|Z}$. As shown in Figure \ref{fig11},  we find that the Sinkhorn GANs also  perform  poorly  in the two relatively simple simulation settings.

The choice  of test statistics in CRT   is crucial  for achieving  adequate statistical power as well as controlling type I errors, whereas it has not been carefully investigated.
For instance,   \citeauthor{bellot2019conditional} \shortcite{bellot2019conditional} proposed to consider multiple test statistics, including  the Maximum Mean Discrepancy (MMD), the Pearson¡¯s correlation coefficient (PCC), the
distance correlation (DC), and the Kolmogorov-Smirnov distance (KS), but little is known about how to appropriately choose test statistics in different scenarios.
 Moreover,   test statistics that solely measure the dependence  between $X$ and $Y$ may  suffer from  inflated type-I errors and/or inadequate power under $H_1$. We consider two scenarios, including (i)
 a simple Markov chain  $X\rightarrow Z\rightarrow Y$ and (ii) $X\rightarrow Z\leftarrow Y$, where
 direct arrows connecting two random variables are direct causes.
 In both scenarios,    test statistics that solely measure the dependence  between $X$ and $Y$  increase type I errors and/or  lose  the statistical power   in testing
 $H_0$ against $H_1$.
Thus,  it is  required to use test statistics that can
  capture the
conditional dependence.
In this paper, we consider  the conditional mutual information (CMI) for $(X, Y, Z)$, denoted as $I(X;Y|Z)$,  and its empirical version \cite{mukherjee2020ccmi}, since it   provides a strong theoretical guarantee for conditional dependence relations such that  $I(X;Y|Z)=0\iff X \perp \!\!\! \perp Y|Z$ \cite{cover2012elements}. The  CMI  has been  widely used in  causal learning \cite{hlinka2013reliability}, graph models \cite{liang2008gene}, and feature selection \cite{fleuret2004fast}. 
However,  the empirical CMI can be computationally difficult especially for high dimensional $Z$.


In this paper, we propose   a novel CIT method  based on the 1-nearest-neighbor sampling strategy (NNSCIT)
  to simulate samples from a distribution that is  approximately close to the true density $p_{X|Z}$.  The nearest-neighbor sampling  first developed by \citeauthor{Fix1951} \shortcite{Fix1951} has been  widely used in density estimation, classification,  and regression problems \cite{silverman2018density,cover1967nearest,devroye1994strong}. Recently,
\citeauthor{sen2017model} \shortcite{sen2017model} used the nearest-neighbor bootstrap procedure to generate samples from  the joint distribution of $(X, Y, Z)$ under $X \perp \!\!\! \perp Y|Z$.  Compared with GANs,  1-nearest-neighbor (1-NN)  not only demonstrates  computational efficiency,  but also exhibits superiority in  approximating quality.

We make four major contributions as follows.  First, we propose to use the  1-NN method to generate samples from the  approximated  conditional distribution of $X$ given $Z$. The 1-NN is computationally much more efficient than  WGANs \cite{bellot2019conditional} and the Sinkhorn GANs \cite{shi2021double}. Theoretically, we show that the distribution of  samples generated from 1-NN is very close to the true conditional distribution in terms of the total variation distance. Second, we take $I(X;Y|Z)$ as our test statistic and estimate it empirically using the recent classifier-based method \cite{mukherjee2020ccmi}. Third, for the  pseudo samples $\widetilde{X}^{(m)}$  ($m=1,\ldots, M$) generated from 1-NN,  we provide  insights to replace $I(\widetilde{X}^{(m)};Y|Z)$ with $I(\widetilde{X}^{(m)};Y)$  to speed up the  calculation, { because estimations of $I(\widetilde{X}^{(m)};Y|Z)$s  are very computationally intensive especially for the case that the  dimensionality of $Z$ is high.} Fourth,  our proposed test   not only asymptotically achieves a valid control of the type I error, but also outperforms all competing tests in numerical studies.

\section{1-Nearest-Neighbor Sampling}
In this section, we present the 1-NN sampling algorithm, as well as its theoretical and empirical results stating that  the distribution of
the sample generated   resembles closely the true conditional distribution.

\subsection{1-NN Sampling from $p_{X|Z}(x|z)$}

We have two data sets  $V_1$ and  $V_2$, both with sample size $n$, such that $V=V_1\cup V_2$ consisting of $2n$ i.i.d. samples from the distribution $p_{X,Z}(x,z)$. Given all $Z$ coordinates in $V_2$,  Algorithm \ref{NNS} presents the procedure to generate a data set $U_0$ consisting of $n$ samples, which mimics  samples generated from $p_{X|Z}(x|z)$. Specifically,  for each $Z$ coordinate in $V_2$, we search the nearest neighbor $(\widetilde{X},\widetilde{Z})$ in $V_1$ in terms of the $Z$ coordinate in $l_2$ norm and then add $\widetilde{X}$ to $U_0$. When $V$ is a set containing samples from the distribution $p_{X,Y,Z}(x,y,z)$, Algorithm \ref{NNS} continues to work with the $Y$-coordinates ignored.



\begin{algorithm}[tb]
\caption{1-Nearest-Neighbor sampling (1-NN($V_1$,$V_2$,$n$))}
\label{NNS}
\textbf{Input}: Data sets $V_1$ and $V_2$, both with sample size $n$ and $V=V_1\cup V_2$ consists of $2n$ i.i.d. samples from $p_{X,Z}$.\\
\textbf{Output}: Generate  $\widetilde{X}$ from $X|Z$ for each $Z$-coordinate in $V_2$.
\begin{algorithmic}[1] 
\STATE Let $U_0=\emptyset$.
\FOR{$(X,Z)$ in $V_2$}
\STATE Go to $V_1$ to find the sample $(\widetilde{X}, \widetilde{Z})$ such that  $\widetilde{Z}$ is  the 1-nearest neighbor of $Z$   in terms of the $l_2$ norm.
 \STATE  $U_0=U_0\cup \{ \widetilde{X} \}$.
\ENDFOR
\STATE \textbf{return} $U_0$
\end{algorithmic}
\end{algorithm}

\subsection{Theoretical Results}

For a given $Z$ coordinate in $V_2$, we show  that the distribution of    $\widetilde{X}$ generated in Algorithm \ref{NNS} is very close to the true conditional distribution in terms of the total variation distance. Before presenting our theoretical result, we first introduce   Lemma 1 of \citeauthor{cover1967nearest} \shortcite{cover1967nearest}, which states  that  the nearest neighbor of $Z$ converges almost surely to $Z$ as the training size $n$ grows to infinity.


 \begin{lem}
 Let $Z$ and $Z_1, Z_2,\ldots, Z_n$ be i.i.d. random variables according to $p(z)$. Let $Z'_{n}$ be the nearest neighbor to $Z$ from the set $\{Z_1,Z_2,\ldots,Z_n\}$. Then $Z'_{n}$  converges almost surely to $Z$ as $n$ grows to infinity. 
 \end{lem}

 We next present several standard regularity conditions, which 
 have
been introduced  in \citeauthor{Weihao2016} \shortcite{Weihao2016},  \citeauthor{Weihao2017} \shortcite{Weihao2017} and  \citeauthor{sen2017model} \shortcite{sen2017model}.  For  the sake of simplicity,  subscripts may be dropped. For example,
 $p(x|z)$ may be used in place of $p_{X|Z}(x|z)$.

\noindent \textbf{Smoothness assumption on $p(x|z)$:}
A smoothness condition is assumed on $p(x|z)$, which can be regarded as a generalization of the boundedness of the maximum eigenvalue of Fisher Information matrix of $x$ w.r.t $z$.  

\noindent{\bf Assumption 1.}  
For all $z\in R ^{d_{z}}$ and all $a$ such that $\left\lVert a-z\right\rVert _{2}\leq \epsilon _1$, we have $0\leq \lambda _{max}(I_{a}(z))\leq \beta $, where $\beta>0$, $\lVert \cdot\rVert_2$ is the
$l_2$ norm and  the generalized curvature matrix $I_{a}(z)=(I_{a}(z)_{ij})$ is defined as
 \begin{eqnarray*}
 \begin{aligned}
  &I_{a}(z)_{ij}=E \left(-\frac{\partial ^{2}\log p(x|\widetilde{z})}{\partial \widetilde{z}_{i}\partial \widetilde{z}_{j}}|_{\widetilde{z}=a}\bigg |Z=z  \right) \\
  &=\left(\frac{\partial ^{2}}{\partial \widetilde{z}_{i}\partial \widetilde{z}_{j}}\int \log \frac{p(x|z)}{p(x|\widetilde{z})}p(x|z)dx  \right)\bigg |_{\widetilde{z}=a}.
\end{aligned}
\end{eqnarray*}

\noindent \textbf{Smoothness assumptions on $p(z)$}: 

\noindent{\bf Assumption 2.} The probability density function $p(z)$ is twice continuously differentiable, and the Hessian matrix $H_{p}(z)$ of the $p.d.f.~p(z)$ with respect to $z$ satisfies $\lVert H_{p}(z)\lVert_{2}\leq c_{d_{z}}$ almost everywhere, where $c_{d_{z}}$ is only dependent on  $d_z$.\\

Given $Z$, let $\widetilde{X}$ denote the sample produced by 1-NN  such that  $\widetilde{X}=X'_n$ is the $X$-coordinate of the sample $(X'_n,Z'_n)$ in $V_1$ with $Z'_n$ being  the nearest neighbor of $Z$. There is no doubt that  $\widetilde{X}\sim p(x|Z'_n)$. Let $\widehat{p}(x|Z):=p(x|Z'_n)$. For any two distributions $P_1$ and $P_2$ that are defined on the same probability space, the total variation distance between $P_1$ and $P_2$ is defined as $d_{TV}(P_1,P_2)=\sup_{A \subset \Omega}|P_1(A)-P_2(A)|$, where the supremum is taken over all measurable subsets of the sample space $\Omega$. We have the following theorem and leave its proof in the Supplementary Materials.
\begin{thm}\label{thm1}
 Under Assumptions 1 and 2,   we have $d_{TV}(p(x|Z),p(x|Z'_n))=d_{TV}(p(x|Z),\widehat{p}(x|Z))=o_{p}(1)$, as the sample size $n$ in $V_1$ goes to infinite.
\end{thm}

 \subsection{Empirical Goodness of Fit }
In this subsection,  we investigate the empirical
goodness-of-fit performance of  samples generated from 1-NN. 
We consider the following two scenarios.

\noindent  {\bf Scenario 1.} $X\sim \mbox{Uniform}[0,1]$ and $Z$ are assumed to be independent, where  $Z$ is a $50$-dimensional multivariate Gaussian distribution with mean vector $(0.7,0.7,\ldots,0.7)$ and  the identity covariance matrix.  The true conditional distribution of $X|Z$ is the same with that of $X$.

\noindent  {\bf Scenario 2.}  Set $X=A_{f}^TZ+\epsilon$, where  the entries of $A_f$  are randomly and uniformly sampled from $[0,1]$ and  then normalized to the unit $l_1$ norm and $Z$ is generated from a $50$ dimensional multivariate Gaussian distribution with mean vector  $(0.7,0.7,\ldots,0.7)$ and the identity covariance matrix.  The noise variables $\epsilon$'s are independently sampled from a normal distribution with mean zero and variance 0.49.

\begin{figure}[t]
\centering
\includegraphics[width=1.1\columnwidth]{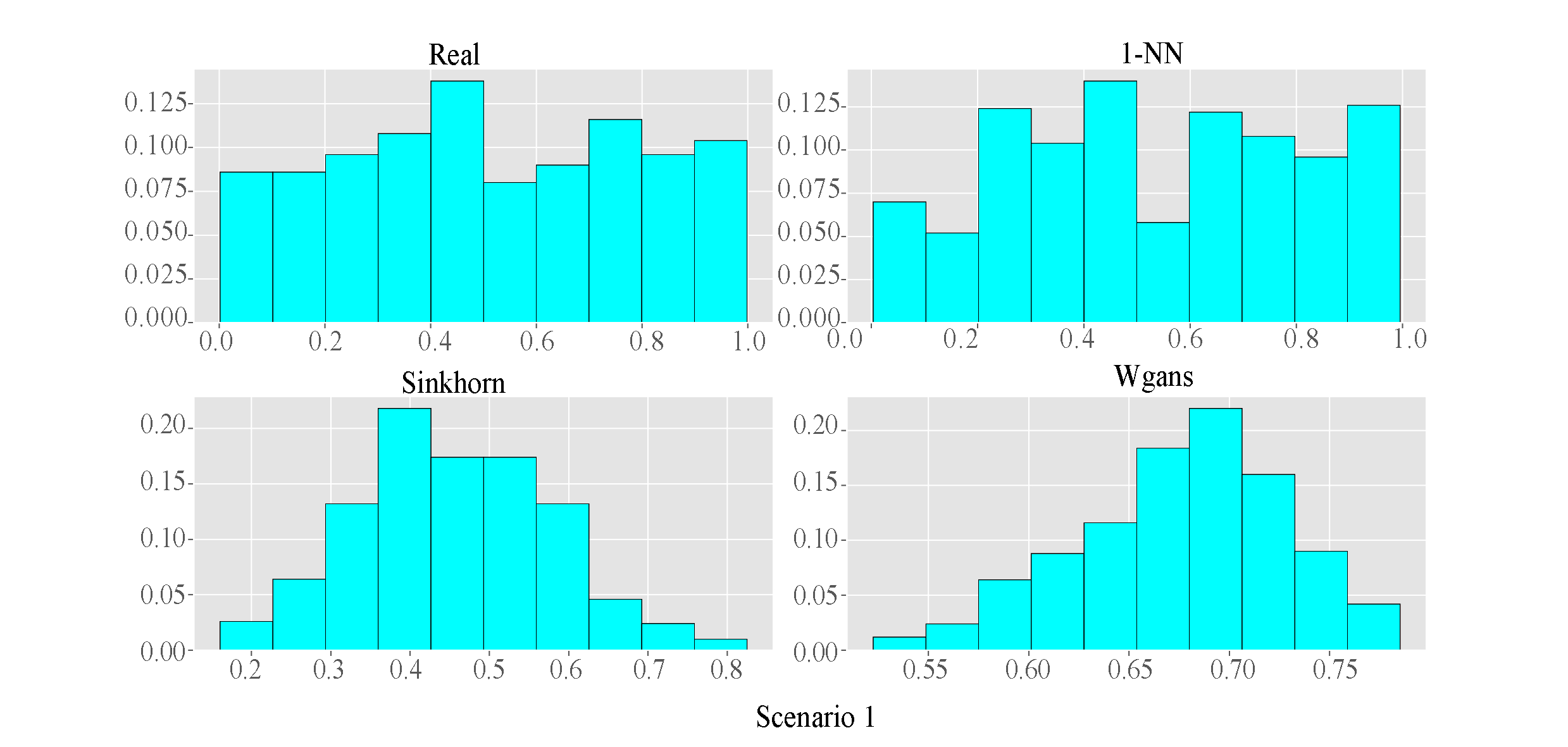}\\
 \includegraphics[width=1.1\columnwidth]{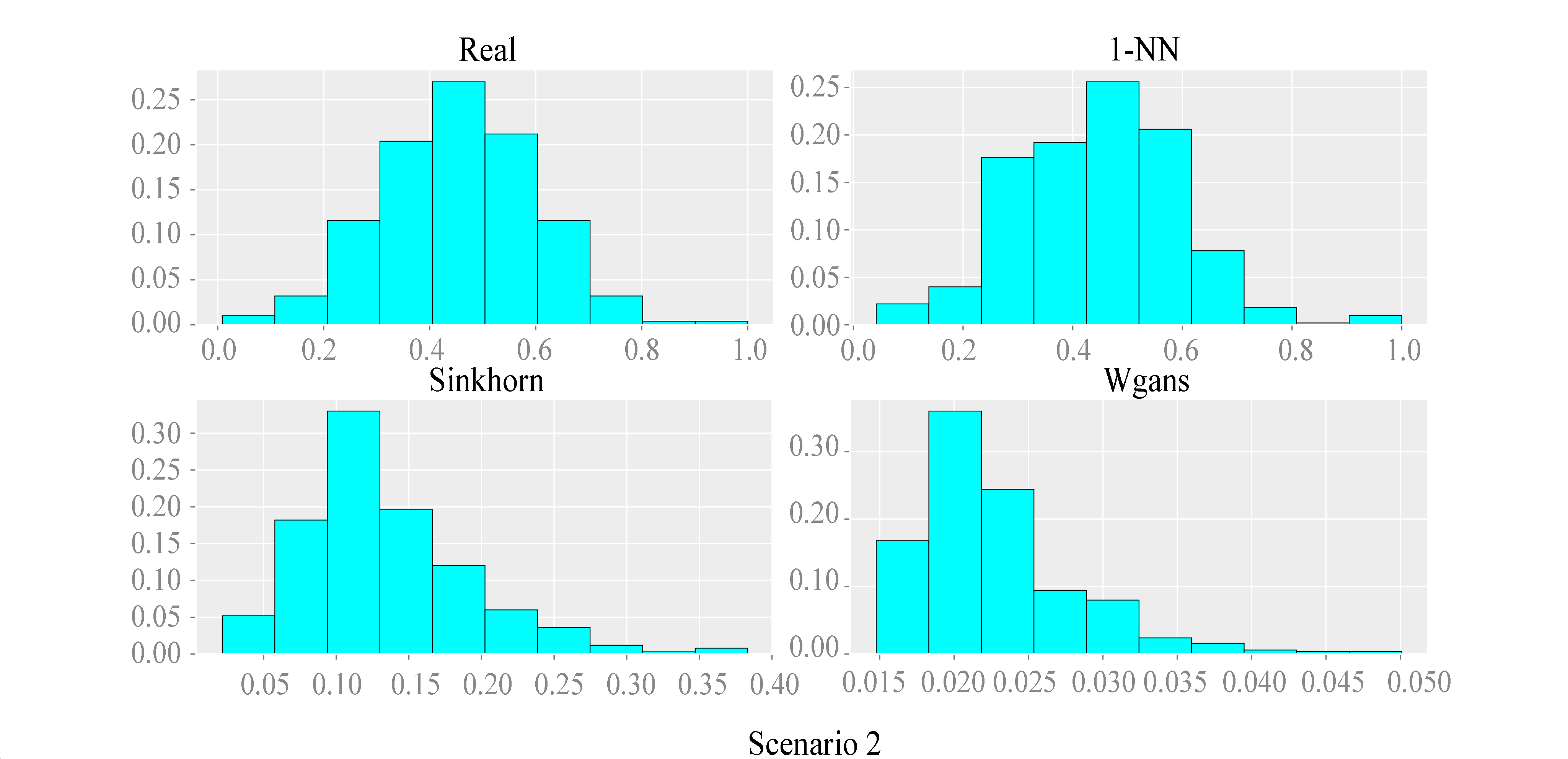}
\caption{\normalsize The conditional histograms. }
\label{fig11}
\end{figure}

For each of the two models, we generate $n=1000$ samples. Randomly choose $500$ samples as the training dataset $V_1$ and the remaining as the testing dataset $V_2$. For 1-NN, we generate $500$ pseudo samples by 1-NN$(V_1,V_2,500).$ Given each $Z$ coordinate in $V_2$, we also generate  pseudo samples $\widetilde{ X}$ using  the WGANs and the Sinkhorn GANs, respectively.  Figure \ref{fig11}  shows the conditional
histograms of the generated samples as well as the true samples all normalized to range $[0,1]$ for Scenarios 1 and 2, respectively. It is observed that   the 1-NNs fit the conditional densities reasonably well, whereas
   WGANs and the Sinkhorn GANs perform poorly. Specifically,   WGANs tend to be biased towards either $0$ or $1$, and the Sinkhorn GANs cannot  capture the feature of the true conditional distribution.

\section{Nearest-Neighbor Sampling Based CIT}

In this section, we  introduce  our CIT based on the nearest-neighbor sampling (NNSCIT) and present the  pseudo code of computing NNSCIT and its $p$-value in Algorithm \ref{main:algorithm}. Moreover,  theoretically, we show that our proposed test achieves a valid control of the type-I error.

\subsection{The Proposed CIT Approach}

Our CIT test is based on an approximation of CMI $I(X; Y|Z)=I(X;Y,Z)-I(X;Z)$, where
$I(X;Y,Z)$ and $I(X;Z)$ are, respectively, the mutual information of $(X;Y,Z)$ and that of $(X;Z)$.
We construct our CIT statistic as a classifier-based CMI estimator (CCMI, \citeauthor{mukherjee2020ccmi}, \citeyear{mukherjee2020ccmi}) of $I(X; Y|Z)$ given by
 \begin{equation}\label{CCMI}
\widehat{I}(X;Y|Z)=\widehat{I}(X;Y,Z)-\widehat{I}(X;Z),
\end{equation}
where $\widehat{I}(X;Y,Z)$ (or $\widehat{I}(X;Z)$) is a classifier-based estimator of  $I(X;Y,Z)$ (or $I(X;Z)$).
By Theorem  1  in \citeauthor{mukherjee2020ccmi} \shortcite{mukherjee2020ccmi}, $\widehat{I}(X;Y|Z)$ is a consistent estimator of ${I}(X;Y|Z)$.
Furthermore,  generate samples $\widetilde{X}^{(m)}$  ($m=1,\ldots, M$) from 1-NN conditioned on   $Z$,
  we can show that  $\widehat{I}(\widetilde{X}^{(m)};Y|Z)-{I}(\widetilde{X}^{(m)}; Y |Z)$   converges to zero  for  all $m$.

Without loss of generality, we discuss how to approximate
 $I(X;Z)=D_{KL}(p_{X,Z}(x,z)||p_X(x)p_Z(z))$, where $p_{X, Z}(x, z)$ is the joint density of $(X, Z)$ and
 $p_X(x)$ and $p_Z(z)$ are, respectively, the marginal density of $X$ and $Z$. Moreover,
  $D_{KL}(F||G)=\int f(x)\log(f(x)/{g(x)}) dx$ is  the Kullback-Leibler (KL) divergence between
  two distribution functions  $F$ and $G$, whose density functions are given by $f(x)$ and $g(x)$, respectively.
The   Donsker-Varadhan (DV) representation of  $D_{KL}(F||G)$ is given by
\begin{equation}\label{DV}
    \sup_{s\in \mathcal{S}} \left[E_{x\sim f}s(x)-\log\{E_{x\sim g}\exp(s(x))\}\right],
\end{equation}
where the function class $\mathcal{S}$ includes all functions with finite expectations in (\ref{DV}).
The optimal function in  (\ref{DV}) is given by $s^{*}(x)=\log({f(x)}/{g(x)})$ \cite{belghazi2018mutual}, leading to
\begin{equation}\label{DVoptimal}
    D_{KL}(F||G)=E_{x\sim f}\log \left\{\frac{f(x)}{g(x)}\right\}-\log\left[E_{x\sim g} \left\{\frac{f(x)}{g(x)}\right\}\right].
\end{equation}
Following  \citeauthor{mukherjee2020ccmi} \shortcite{mukherjee2020ccmi}, we    use the classier two-sample principle \cite{lopez2017revisiting} to  estimate the likelihood ratio $L(x)={f(x)}/{g(x)}$ as follows. Specifically, we consider $n$ i.i.d. samples $\{X_{i}^f \}_{i=1}^{n}$ with $X_{i}^{f}\sim f(x)$ and $d$ i.i.d. samples $\{X_{j}^g \}_{j=1}^{d}$ with $X_{j}^{g}\sim g(x)$.
We  label  $y_i^f=1$ for all $X_i^f$ and $y_j^g=0$ for all $X_j^{g}$.
One trains a binary classifier using deep neural network on this supervised classification task. The classifier produces predicted probability $\alpha_{l}=Pr(y=1|X_l)$ for a given sample $X_l$, leading to  an estimator of the likelihood ratio on $X_l$   given  by $\widehat{L}(X_l)= {\alpha_{l}}/(1-\alpha_{l})$.  Therefore,  it follows from (\ref{DVoptimal}) that
an estimator of the KL-divergence, $\widehat{D}_{KL}(F||G)$, is given by
\begin{equation*}\label{MIE}
{n}^{-1}    \sum _{i=1}^{n}\log \widehat{L}(X_{i}^{f})-\log \left\{{d}^{-1}\sum _{j=1}^{d}\widehat{L}(X_{j}^{g})\right\}.
\end{equation*}
Since mutual information   is a  special case of the KL divergence, we   obtain the estimator $\widehat{I}(X;Z)$ of ${I}(X;Z)$ and that of  ${I}(X;Y,Z)$.

 Following the idea of CRT, the $p$-value of our CIT method can be given by
\begin{equation}\label{CRTCMI}
    p=\frac{1+\sum _{m=1}^{M} I\left(\widehat{I}(\widetilde{X}^{(m)};Y|Z)\geq \widehat{I}(X;Y|Z)\right)}{1+M}.
\end{equation}
In Lemma 3, we   show  that the excess type I error of the test based on (\ref{CRTCMI}) is bounded by the total variation distance between $p_{\bm X|\bm Z}(\cdot|\bm Z)$ and $\widehat{p}_{\bm X|\bm Z}(\cdot|\bm Z)$. By Theorem \ref{thm1}, we further get $P(p\leq \alpha |H_0) \leq \alpha+o(1)$. Therefore, the excess type I error of our CIT method is guaranteed to tend to zero as $n\rightarrow\infty$. Two binary classifications based on deep neural network should be trained to get $\widehat{I}(\widetilde{X}^{(m)};Y|Z)$ for each $m$. Together with $\widehat{I}({X};Y|Z)$, $2(M+1)$ binary classification neural networks should be trained for computing the $p$-value in (\ref{CRTCMI}). When $M$ is large, the calculation is extremely intensive and time consuming, especially for the case that the dimensionality of $Z$ is high.

In (\ref{CRTCMI}), instead of using $\widehat{I}(\widetilde{X}^{(m)};Y|Z)$, we further propose to  utilize   $\widehat{I}(\widetilde{X}^{(m)};Y)$ calculated by the method of \citeauthor{mesner2020conditional} \shortcite{mesner2020conditional} according to the following reasons.  First, compared with $\widehat{I}(\widetilde{X}^{(m)};Y|Z)$, $\widehat{I}(\widetilde{X}^{(m)};Y)$ is computationally very fast. Second, 
 $\widetilde{X}^{(m)}$ is generated from 1-NN conditional on    $Z$,   we thus have $I(\widetilde{X}^{(m)};Y|Z)=0$, whereas  $\widetilde{X}^{(m)}$ and $Y$ may share information via $Z$, that is, $I(\widetilde{X}^{(m)};Y)\geq I(\widetilde{X}^{(m)};Y|Z)=0$. By the consistency of $\widehat{I}(\widetilde{X}^{(m)};Y|Z)$ and $\widehat{I}(\widetilde{X}^{(m)};Y)$, we conclude that replacing $\widehat{I}(\widetilde{X}^{(m)};Y|Z)$ with $\widehat{I}(\widetilde{X}^{(m)};Y)$ can improve controlling  the probability of making type I error of our CIT method. Thus, we  propose a simple counterpart of (\ref{CRTCMI}) for $p$-value calculation as follows:
\begin{equation}\label{CRTCMI2}
  p=\frac{1+\sum _{m=1}^{M} I\left(\widehat{I}(\widetilde{X}^{(m)};Y)\geq \widehat{I}(X;Y|Z)\right)}{1+M}.
\end{equation}
Since $\widetilde{X}^{(m)}_i$s are generated by the 1-NN sampling strategy, we call our test as NNSCIT.
Equation 
(\ref{CRTCMI2}) lays the foundation of our CIT method, whose    pseudo code has been summarized in Algorithm \ref{main:algorithm}.

We describe how to obtain $\widehat{I}(\widetilde{X}^{(m)};Y)$.
Specifically, given i.i.d. samples $\{(\widetilde{X}_i,Y_i)\}_{i=1}^n$ with $(\widetilde{X}_i,Y_i)\sim p_{\widetilde{X},Y}$.  Let $ \rho _{k,i}/2$ be the $l_{\infty}$-distance from point $(\widetilde{X}_i,Y_i)$ to its $k$th nearest neighbor. Define
\begin{equation*}\label{nwi}
    n_{\widetilde{X},i}=|\{\widetilde{X} _{j}:|\widetilde{X} _{i}-\widetilde{X}_{j}| \leq \rho _{k,i}/2~,j\neq i\}|,
\end{equation*}
where $|A|$ is the number of elements in the set $A$.
Similarly,  define $n_{Y,i}$.  For each $i$, we define
\begin{equation*}
    \delta_i=\psi (k)-\psi (n_{\widetilde{X},i})-\psi (n_{Y,i})+\psi (n),
\end{equation*}
  where   $\psi (x):=d\log\Gamma(x)/dx$ is the digamma function. Therefore, we have
 \begin{equation}\label{MI}
    \widehat{I}(\widetilde{X};Y)=\max\left\{\frac{1}{n}\sum_{i=1}^n\delta_i,0\right\}.
\end{equation}
It follows from  Theorems 3.1 and 3.2 in \citeauthor{mesner2020conditional} \shortcite{mesner2020conditional} that $\widehat{I}(\widetilde{X};Y)$ is a consistent estimator of ${I}(\widetilde{X};Y)$.  

Finally, we discuss why we cannot replace $\widehat{I}(X;Y|Z)$ by $\widehat{I}(X;Y)$ in (\ref{CRTCMI2}).
One may think of approximating $p$-value as follows:
\begin{equation}\label{CRTMI}
  p=\frac{1+\sum _{m=1}^{M} I\left\{\widehat{I}(\widetilde{X}^{(m)};Y)\geq \widehat{I}(X;Y)\right\}}{1+M},
\end{equation}
which results in  another CRT test.
Let $\widehat{c}_{\alpha}$ be the upper $\alpha$ quantile of the distribution of $\widehat{I}(\widetilde{X}^{(m)};Y)$. Given significance level $\alpha$, the rejection regions of (\ref{CRTCMI2}) and (\ref{CRTMI}) are given by $\{\widehat{I}(X;Y|Z)>\widehat{c}_{\alpha}\}$ and $\{\widehat{I}(X;Y)>\widehat{c}_{\alpha}\}$, respectively. Under $H_1$, $I(X; Y|Z)$ should deviate from zero. Intuitively, the test with rejection region $\{\widehat{I}(X;Y|Z)>\widehat{c}_{\alpha}\}$ is more likely to accept $H_1$ than that with $\{\widehat{I}(X;Y)>\widehat{c}_{\alpha}\}$. For example, consider $X\rightarrow Z\leftarrow Y$. This relation indicates $X \not \! \perp \!\!\! \perp Y|Z$ ($H_1$ holds), but $X$ and $Y$ may be independent. Therefore, the rejection region $\{\widehat{I}(X;Y|Z)>\widehat{c}_{\alpha}\}$ could detect $H_1$, but $\{\widehat{I}(X;Y)>\widehat{c}_{\alpha}\}$ may fail to do so. That is, the test using (\ref{CRTCMI2}) is generally more powerful than that using (\ref{CRTMI}) under $H_1$. Consider another special case when $X\perp \!\!\! \perp Z$, we obtain $I(X;Y|Z)\geq I(X;Y)$. By the consistency of $\widehat{I}(X;Y|Z)$
and $\widehat{I}(X;Y)$, replacing $\widehat{I}(X;Y)$ in (\ref{CRTMI}) with $\widehat{I}(X;Y|Z)$ will increase the power under $H_1$. That is, (\ref{CRTCMI2}) endows more power than (\ref{CRTMI}) under $H_1$. We can reach the same conclusion for $Y\perp \!\!\! \perp Z$.

\begin{algorithm}[tb]
\caption{ Nearest-Neighbor sampling based conditional independence test (NNSCIT)}
\label{main:algorithm}
\textbf{Input}: Data-set $U$ of $n$ i.i.d. samples from $p_{X,Y,Z}$.\\
\textbf{Parameter}: The number of repetitions $M$; the neighbor order $k$ in MI estimation;  the significance level $\alpha$.\\
\textbf{Output}:  Accept $H_0:X \perp \!\!\! \perp Y|Z$ or $H_1:X \not \! \perp \!\!\! \perp Y|Z$.
\begin{algorithmic}[1] 
\STATE Randomly divide $U$ into two disjoint parts:
$U_1:=\{X_{train}, Y_{train}, Z_{train} \}$ with sample size  $n-\lfloor n/3 \rfloor$ and $U_2:=\{X_{test}, Y_{test}, Z_{test} \}$ with sample size  $\lfloor n/3 \rfloor$.
\STATE $m=1$.
\WHILE{$m\leq M$}
\STATE Randomly taking $\lfloor n/3 \rfloor$ samples  from $U_1$ to obtain $V_1$.
\STATE Produce $U_0^m:=\{\widetilde{X}^{(m)}\}$ using 1-NN($V_1$,$U_2$, $\lfloor n/3 \rfloor$) in Algorithm \ref{NNS}.
\STATE Compute  $I^{(m)}:=\widehat{I}(\{\widetilde{X}^{(m)}\};\{Y_{test}\})$ according to Equ. (\ref{MI}).
\STATE $m=m+1$.
\ENDWHILE
\STATE Compute  $I:=\widehat{I}(\{X_{test}\};\{Y_{test}\}|\{Z_{test}\})$ according to Equ. (\ref{CCMI}).
\STATE Compute $p$-value: $p:=\frac{1+\sum_{m = 1}^{M}I\{I^{(m)}\geq I\}  }{1+M} $.
\IF {$p\geq \alpha$}
\STATE Accept $H_0:X \perp \!\!\! \perp Y|Z$.
\ELSE
\STATE Accept $H_1:X \not \! \perp \!\!\! \perp Y|Z$.
\ENDIF
\end{algorithmic}
\end{algorithm}

\begin{figure*}[t]
\centering
\includegraphics[width=0.783\textwidth]{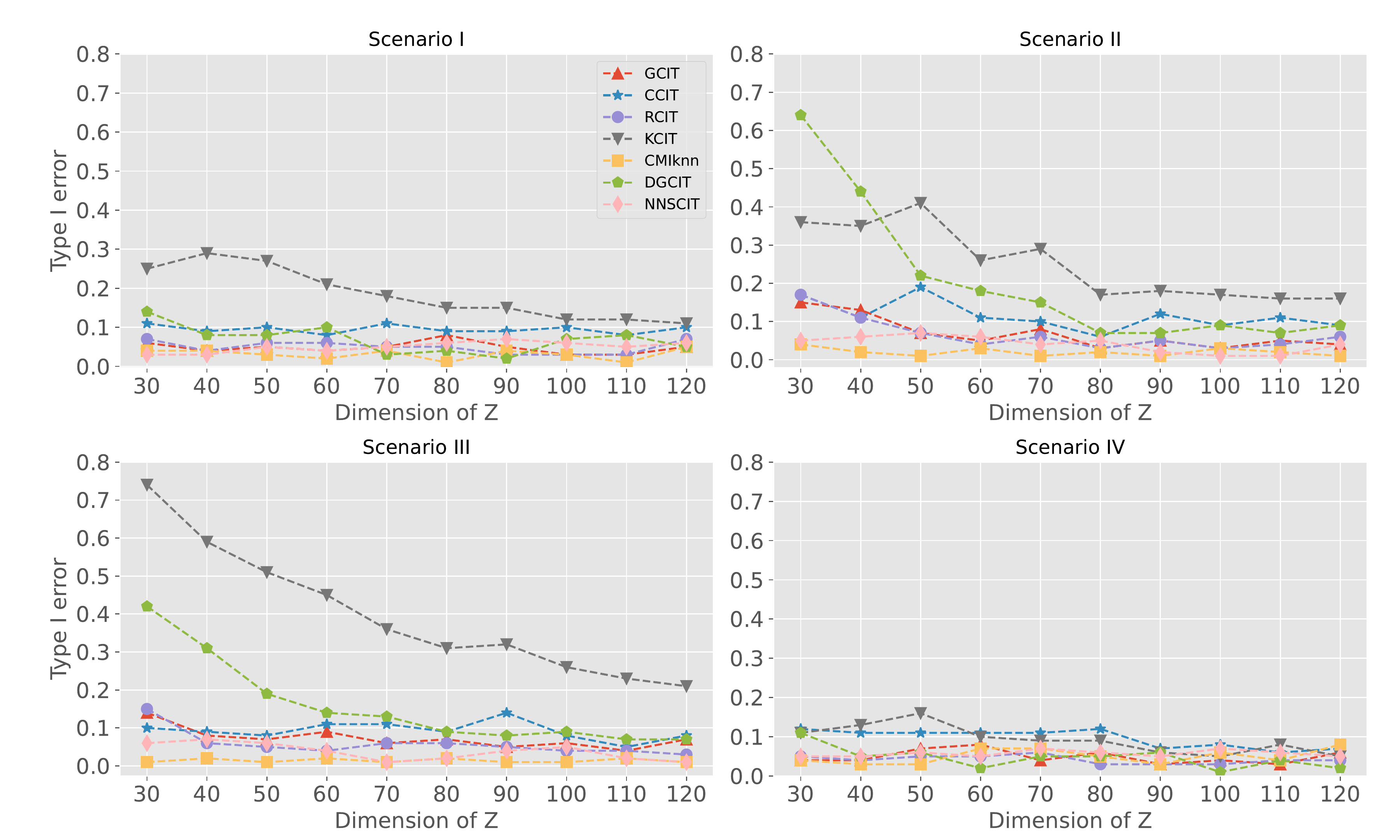}  
\caption{\normalsize The empirical type-I error rate of various tests under  $H_0$.} 
\label{fig22}
\end{figure*}

\begin{figure*}[t]
\centering
\includegraphics[width=0.783\textwidth]{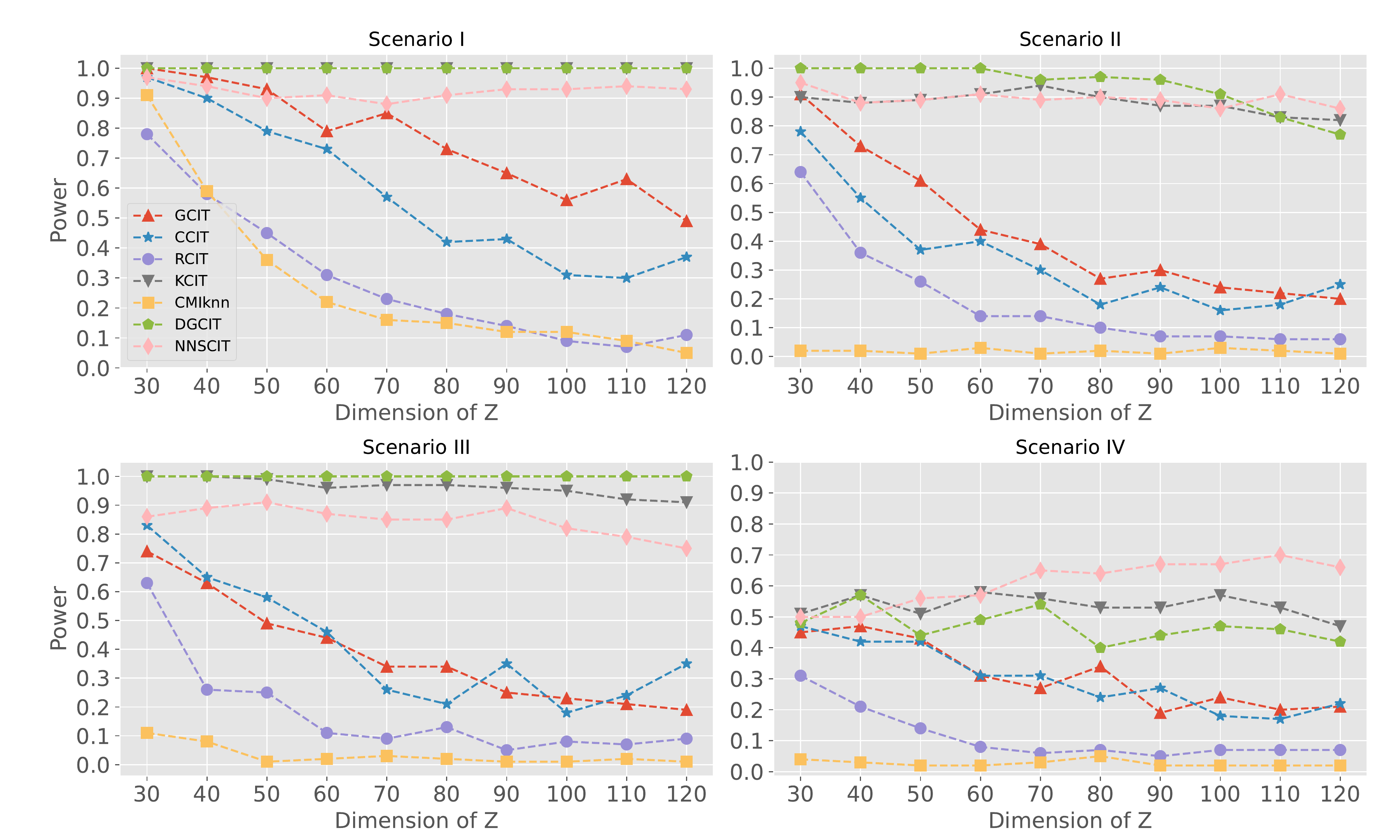}
\caption{\normalsize The empirical power of various tests under $H_1$.} 
\label{fig33}
\end{figure*}

\begin{table*}[t]
\centering
\begin{tabular}{c|c|c|c|c|c|c|c}
$d_Z$&GCIT&CCIT&RCIT&KCIT&CMIknn&DGCIT&NNSCIT \\
 \hline
 5&0.11&0.36&0.53&0.72&0.03&0.50&{0.01} \\
10&0.48&0.22&0.82&0.84&0.06&0.86&{0.05} \\
15&0.87&0.21&0.85&0.93&0.15&0.91&{0.05} \\
20&0.93&0.25&0.90&0.98&0.05&0.97&0.07
\end{tabular}
\caption{The empirical type-I error rate of various tests for Example 1.}
\label{table1}
\end{table*}

\begin{table*}[t]
\centering
\begin{tabular}{c|c|c|c|c|c|c|c}
$d_Z$&GCIT&CCIT&RCIT&KCIT&CMIknn&DGCIT&NNSCIT \\
 \hline
5 &0.37&1&0.03&0.02&1&0.78&1 \\
10&0.53&1&0.04&0.10&1&0.82&1 \\
15&0.55&1&0.05&0.14&1&0.79&1 \\
20&0.63&1&0.07&0.16&1&0.88&1
\end{tabular}
\caption{The empirical power of various tests  for Example 2.}
\label{table2}
\end{table*}

\subsection{Theoretical Results}

In this subsection, we present   theoretical results of our NNSCIT based on (\ref{CRTCMI}) and  (\ref{CRTCMI2}). We introduce the following notation. 
Without loss of generality, let $U_2:=\{(X_1,Y_1,Z_1),\ldots,(X_{n_1},Y_{n_1},Z_{n_1}) \}$ in Algorithm \ref{main:algorithm} with $n_1=\lfloor n/3 \rfloor$,  where $\lfloor x\rfloor$ is the largest integer not greater than $x$. We define $\bm{X}:=(X_1,X_2,\ldots,X_{n_1})$, $\bm{Y}:=(Y_1,Y_2,\ldots,Y_{n_1})$, and $\bm{Z}:=(Z_1,Z_2,\ldots,Z_{n_1})$.
Denote ${P}(\cdot|\bm Z):={p}(\cdot|Z_1)\times\ldots\times{p}(\cdot|Z_{n_1})$ and $\widehat{P}(\cdot|\bm Z):=\widehat{p}(\cdot|Z_1)\times\ldots\times\widehat{p}(\cdot|Z_{n_1})$. Assume that $\widetilde{\bm{X}}^{(m)}:=(\widetilde{X}_{1}^{(m)}, \ldots, \widetilde{X}_{n_1}^{(m)})$ is sampled according to $\widehat{P}(\cdot|\bm Z)$ for $m=1,\ldots, M$.  Let $T(\bm X,\bm Y,\bm Z):=\widehat{I}(X;Y|Z)$ and $T(\bm \widetilde{\bm X}^{(1)},\bm Y,\bm Z):=\widehat{I}(\widetilde{ X}^{(1)};Y|Z), \ldots,$  $T(\bm \widetilde{\bm X}^{(M)},\bm Y,\bm Z):=\widehat{I}(\widetilde{ X}^{(M)};Y|Z)$.

Let $\widetilde{\bm X}_{F}$ be an additional copy sampled from $\widehat{P}(\cdot|\bm Z)$ and
independently of $\bm Y$ and of $\bm X, \widetilde{\bm X}^{(1)}, \ldots, \widetilde{\bm X}^{(M)}$. Under  $H_0$: $X \perp \!\!\! \perp Y|Z$,
conditionally on $\bm Y$ and $\bm Z$,
$\bm X$ and $(\widetilde{\bm X}^{(1)},\ldots, \widetilde{\bm X}^{(M)})$ are independent, and $\widetilde{\bm X}_{F}$  and $(\widetilde{\bm X}^{(1)},\ldots,\widetilde{\bm X}^{(M)})$ are independent. Thus, we have
\begin{align*}
&d_{TV}\{(({\bm X},\widetilde{\bm X}^{(1)},\ldots,\widetilde{\bm X}^{(M)})|\bm Y,\bm Z),\\
&~~~~~~~~((\widetilde{\bm X}_{F},\widetilde{\bm X}^{(1)},\ldots,\widetilde{\bm X}^{(M)})|\bm Y,\bm Z)\} \\
&=d_{TV}\{({\bm X}|\bm Y,\bm Z),(\widetilde{\bm X}_{F}|\bm Y,\bm Z) \}=d_{TV}\{P(\cdot|\bm Z),\widehat{P}(\cdot|\bm Z)\}.
\end{align*}
Define a set $\mathcal{A}_{\alpha }$ as
\begin{align*}
\mathcal{A}_{\alpha }:=&\Big\{(\bm x,\widetilde{\bm x}^{(1)},\ldots,\widetilde{\bm x}^{(M)}):\\
&\frac{1+\sum_{m = 1}^{M}I\{ T(\widetilde{\bm x}^{(m)},\bm Y,\bm Z)\geq T(\bm x,\bm Y,\bm Z) \}   }{1+M} \leq \alpha  \Big\}.
\end{align*}
Then, we have
\begin{align*}
&P(p\leq \alpha |\bm Y, \bm Z)=P\{({\bm X},\widetilde{\bm X}^{(1)},\ldots,\widetilde{\bm X}^{(M)})\in \mathcal{A}_{\alpha }|\bm Y,\bm Z\}\\
&\leq d_{TV}\{(({\bm X},\widetilde{\bm X}^{(1)},\ldots,\widetilde{\bm X}^{(M)})|\bm Y,\bm Z),\\
&~~~~~~~~~~~~~~~~((\widetilde{\bm X}_{F},\widetilde{\bm X}^{(1)},\ldots,\widetilde{\bm X}^{(M)})|\bm Y,\bm Z)\} \\
&~~+P\{(\widetilde{\bm X}_{F},\widetilde{\bm X}^{(1)},\ldots,\widetilde{X}^{(M)})\in \mathcal{A}_{\alpha }|\bm Y,\bm Z\}\\
&=d_{TV}\{P(\cdot|\bm Z),\widehat{P}(\cdot|\bm Z) \}\\
&~~+P\{(\widetilde{\bm X}_{F},\widetilde{\bm X}^{(1)},\ldots,\widetilde{\bm X}^{(M)})\in \mathcal{A}_{\alpha }|\bm Y,\bm Z\}.
\end{align*}
Conditioning on $\bm Y$ and $\bm Z$, $\widetilde{\bm X}_{F},\widetilde{\bm X}^{(1)},\ldots,\widetilde{\bm X}^{(M)}$ are identically distributed and thus  exchangeable, so
$P\{(\widetilde{\bm X}_{F},\widetilde{\bm X}^{(1)},\ldots,\widetilde{\bm X}^{(M)})\in \mathcal{A}_{\alpha }|\bm Y,\bm Z\}\leq \alpha $ holds and we   obtain the following result.

\begin{lem}\label{thm2}
Assume that $H_0:X \perp \!\!\! \perp Y|Z$ is true, 
for any desired significance level $\alpha \in (0,1)$, the   type I error of  test (\ref{CRTCMI})  satisfies
\begin{equation}\label{type1}
    P(p\leq \alpha |\bm Y, \bm Z) \leq \alpha+d_{TV}\{P(\cdot|\bm Z),\widehat{P}(\cdot|\bm Z)\}.
\end{equation}
\end{lem}
An immediate implication of
 Lemma \ref{thm2} is that  the type I error rate holds unconditionally as follows:
\begin{equation*}
    P(p\leq \alpha |H_0) \leq \alpha+E[d_{TV}\{P(\cdot|\bm Z),\widehat{P}(\cdot|\bm Z)\}].
\end{equation*}
Furthermore, for any given test statistic $T(\cdots)$, we can compute the $p$-value  via (\ref{CRT}) by replacing $\bm{X}^{(m)}$ with the 1-NN sample $\widetilde{\bm{X}}^{(m)}$. The resulting test also  enjoys (\ref{type1}) by  similar arguments.

Under $H_0$, $I(\widetilde{X}^{(m)};Y)\geq I(\widetilde{X}^{(m)};Y|Z)$. Denote the $p$ values in (\ref{CRTCMI}) and (\ref{CRTCMI2}) as $p$ and $p^*$, respectively. With the consistency of $\widehat{I}(\widetilde{X}^{(m)};Y|Z)$ and $\widehat{I}(\widetilde{X}^{(m)};Y)$,
 we obtain the following main result.

\begin{thm}\label{thm3}
Assume that $H_0$ holds, we have 
\begin{align*}
P(p^*\leq \alpha |H_0)-\alpha  &\leq   P(p\leq \alpha |H_0)-\alpha\\
& \leq E[d_{TV}\{P(\cdot|\bm Z),\widehat{P}(\cdot|\bm Z)\}].
\end{align*}
\end{thm}
Theorem \ref{thm3} has three important implications. First,  the excess  type I error over a desired level $\alpha\in (0,1)$ of the test (\ref{CRTCMI2})  is  bounded by $E\{d_{TV}(\widehat{P}(\cdot|\bm Z),P(\cdot|\bm Z))\}$. Second, our proposed method outperforms CRT (\ref{CRTCMI}) in controlling type I error. Third,
by Theorem \ref{thm1},  we  get $$P(p^*\leq \alpha |H_0) \leq \alpha+o(1).$$ Thus, the excess type I error of our NNSCIT is guaranteed to be small.

\section{Performance Evaluation}
In this section,  we examine the finite sample performance  of our  NNSCIT by using the synthetic datasets.  We compare NNSCIT  with   GCIT  \cite{bellot2019conditional}, the classifier-based CI test (CCIT)  \cite{sen2017model}, the kernel-based CI test (KCIT) \cite{zhang2011kernel},  RCIT \cite{strobl2019approximate}, the CMI-based CI test (CMIknn) \cite{runge2018conditional}, and  DGCIT  \cite{shi2021double}. We leave some additional simulation studies and the real data analysis  in the Supplementary Materials.
The source code of NNSCIT is available at \url{https://github.com/LeeShuai-kenwitch/NNSCIT}.

\subsection{Performances on Synthetic Dataset}

 The synthetic data sets are generated by using the post non-linear model similar to those in \citeauthor{zhang2011kernel} \shortcite{zhang2011kernel}; \citeauthor{doran2014permutation}  \shortcite{doran2014permutation}; and \citeauthor{bellot2019conditional} \shortcite{bellot2019conditional}. Specifically, we define $(X,Y,Z)$ under $H_0$ and $H_1$ as follows:
\begin{eqnarray*}
    && H_0:X=f(A_{f}^TZ+\epsilon_{f}),~Y=g(A_{g}^TZ+\epsilon_{g}), \nonumber \\
    && H_1:Y=h(A_{h}^TZ+b X)+\epsilon_{h}.
\end{eqnarray*}
The entries of $A_f$ and $A_g$ are randomly and uniformly sampled from $[0,1]$ and then normalized to the unit $l_1$
 norm. The entries of $A_h$ are sampled from a standard normal distribution and   $b$ is set to  $2$. The noise variables $\epsilon_f$, $\epsilon_g$ and $\epsilon_h$ are independently sampled from a normal distribution with mean zero and variance 0.49. The significance level is set at $\alpha=0.05$ and the sample size is fixed at $n=1000$. Set $M=500$ and $k=3$. Consider the following four scenarios:

\noindent \textbf{ Scenario I.} Set $f$, $g$ and $h$ to be the identity functions,  inducing linear dependencies, $Z\sim N(0.7,1)$, and $X\sim N(0,1)$ under $H_1$.

\noindent \textbf{ Scenario II.}\ Set $f$, $g$ and $h$ as in Scenario I, but use a Laplace distribution to generate $Z$.

\noindent \textbf{ Scenario III.} Set $f$, $g$ and $h$ as in Scenario I, but use Uniform$[-2.5,2.5]$  to generate $Z$.

\noindent \textbf{ Scenario IV.}  Set $f$, $g$ and $h$ to be randomly sampled from $\left\{ x^{2}, x^{3}, \mbox{tanh}(x), \cos(x)\right\}$. Set $Z\sim N(0,1)$, and $X\sim N(0,1)$ under $H_1$.

We vary the dimension of $Z$ as $d_Z=30,$ $40,$ $50,$ $60,$ $70,$ $80,$ $90,$ $100,$ $110,$ and  $120$. Figures \ref{fig22} and \ref{fig33} include the type-I error rates under $H_0$ and powers under $H_1$, respectively, over $300$ data replications. Additional simulation results for $d_Z=5, 10, 15, 20,$ and $25$ can be found in the Supplementary Materials  (Figures 1  and 2).

We have the following observations. First, our test controls  type I error very well under $H_0$, while achieves high power under $H_1$. Second,  CMIknn has satisfactory performances in controlling  type-I error, but under $H_1$, it loses power  in almost all scenarios. Third,  although DGCIT and KCIT have adequate power under $H_1$,  they  have inflated type-I errors in some cases, especially when $d_Z$ is less than $30$. Fourth,
GCIT, CCIT and  RCIT  cannot control type-I errors  in some cases,  especially when $d_Z$ is less than $30$. Moreover, under $H_1$, GCIT, CCIT and  RCIT lose some power in almost all scenarios.

Figure 4  in the Supplementary Materials reports  the run times as a function of    $d_Z$ for a single CIT with data generated under Scenario II.  Other scenarios show similar performance.
 Our NNSCIT is  computationally very efficient. In contrast, CCIT, CMIknn and DGCIT are very time-consuming and are prohibitive in practice.

\subsection{Performances on Two Examples}
As discussed in the Introduction, we evaluate  the performances of our method in the following two examples. The details of data generation mechanisms are presented in the Supplementary Materials.

\noindent \textbf{Example 1.}  $X\rightarrow Z\rightarrow Y$. In this case, $H_0$ holds, but there is a strong dependence between $X$ and $Y$. Table \ref{table1} reports the type-I error rates. Our NNSCIT controls type-I error very well, but GCIT, CCIT, RCIT, KCIT and DGCIT break down as their type-I errors are very large. 

\noindent \textbf{Example 2.}  $X\rightarrow Z\leftarrow Y$. In this case, $H_1$ holds, but $X$ and $Y$ are independent.  Table  \ref{table2} reports the powers of different methods. Our method achieves  power as high as $1$. In contrast, RCIT and  KCIT have power less than $0.2$ and  GCIT and DGCIT also lose some power.

\section{Conclusion}
In this paper, we propose a novel and fast   NNSCIT. We use the 1-NN sampling strategy  to approximate the conditional distribution  $X|Z$. Compared with
GANs, 1-NN not only has
computational efficiency, but also exhibits advantage in approximation accuracy. We take the
classifier-based conditional mutual information (CCMI) estimator as
our test statistic, which     captures the
conditional-dependence feature very well. We show that
our   NNSCIT has three notable features, including controlling type-I error well,  achieving high power under $H_1$, and being  computationally efficient.

\section{Acknowledgments}
Dr. Ziqi Chen's work was partially supported by National Natural Science
Foundation of China (NSFC) (12271167 and 11871477) and  Natural Science Foundation of
Shanghai (21ZR1418800).
Dr Christina Dan Wang's work was partially supported by National Natural Science
Foundation of China (NSFC) (12271363 and 11901395).

\bibliography{aaai23_mod}
\end{document}